\documentclass[10pt,twocolumn,letterpaper]{article}

\usepackage{wacv}

\usepackage{booktabs}
\usepackage{pgfplotstable}
\pgfplotsset{compat=1.16}
\usepackage{array}
\usepackage{makecell} %

\pgfplotstableset{
  col sep=comma,
  header=has colnames,      %
  trim cells=true,
  fixed, 
  precision=2, 
  fixed zerofill,  %
  string type,                         %
  every head row/.style={before row=\toprule, after row=\midrule},
  every last row/.style={after row=\bottomrule},
}

\pgfplotstableread[col sep=comma]{./data/DSMIII_tertile_ranges.csv}\cielab

\newcommand{\getcellidxp}[4][1]{%
  \pgfplotstablegetelem{#2}{[index]#3}\of\cielab
  \begingroup
    \pgfkeys{/pgf/number format/.cd, fixed, precision=#1, zerofill}%
    \pgfmathprintnumberto{\pgfplotsretval}{\getcelltmp}%
    \xdef#4{\getcelltmp}%
  \endgroup
}

\getcellidxp[1]{0}{1}{\TOneL}
\getcellidxp[1]{0}{2}{\TOneH} 
\getcellidxp[1]{1}{1}{\TTwoL} 
\getcellidxp[1]{1}{2}{\TTwoH} 
\getcellidxp[1]{2}{1}{\TThrL}
\getcellidxp[1]{2}{2}{\TThrH}

\definecolor{wacvblue}{rgb}{0.21,0.49,0.74}
\usepackage[pagebackref,breaklinks,colorlinks,allcolors=wacvblue]{hyperref}

\def\wacvPaperID{4} %
\def\confName{WACV}
\def\confYear{2026}

\title{Robustness of Presentation Attack Detection in Remote Identity Validation Scenarios}

\author{
John J. Howard\textsuperscript{1,+} \quad
Richard O. Plesh\textsuperscript{1,*,+} \quad
Yevgeniy B. Sirotin\textsuperscript{1,+} \quad
Jerry L. Tipton\textsuperscript{1,+} \\
\textsuperscript{1}The SAIC Identity and Data Sciences Laboratory\\
\textsuperscript{*}\small Corresponding author: {\tt\small rplesh@idslabs.org}.\\
\textsuperscript{+}\small Authors listed alphabetically.
\\\\
Arun R. Vemury\textsuperscript{2}\\
\textsuperscript{2}The United States Department of Homeland Security, Science and Technology Directorate
}

\begin{document}
\maketitle
\begin{abstract}
Presentation attack detection (PAD) subsystems are an important part of effective and user-friendly remote identity validation (RIV) systems. However, ensuring robust performance across diverse environmental and procedural conditions remains a critical challenge. This paper investigates the impact of low-light conditions and automated image acquisition on the robustness of commercial PAD systems using a scenario test of RIV. Our results show that PAD systems experience a significant decline in performance when utilized in low-light or auto-capture scenarios, with a model-predicted increase in error rates by a factor of about four under low-light conditions and a doubling of those odds under auto-capture workflows. Specifically, only one of the tested systems was robust to these perturbations, maintaining a maximum bona fide presentation classification error rate (BPCER\textsubscript{max}) below 3\% across all scenarios. Our findings emphasize the importance of testing across diverse environments to ensure robust and reliable PAD performance in real-world applications.
\end{abstract}%

\section{Introduction}
\label{sec:intro}%
Remote Identity Validation (RIV) has become an essential technology for convenient identity verification, particularly in digital interactions. With the increasing reliance on digital services, RIV enables individuals to verify their identities online in lieu of in-person interactions, improving accessibility and efficiency for various use-cases such as account-creation for banking and social media or requests for government services. However, ensuring the robustness and usability of RIV systems remains a critical challenge, particularly in the presence of environmental and procedural variations that may impact their effectiveness.
\pgfplotstableread{./data/bpcer_stats.csv}\bpcerstats
\begin{table}[t]
  \centering
  \caption{Descriptive statistics of BPCER per scenario, pooled across PAD subsystems and smartphone configurations. Bold values indicate when the 3\% threshold has been successfully met.}
  \label{tab:bpcerstats}
  \begingroup
    \setlength{\tabcolsep}{2.5pt}
    \pgfplotstabletypeset[
      columns/{Metric}/.style={
        column name={\makecell[l]{Metric}},
        column type=l
      },%
      columns/{low-light}/.style={
        column name={\makecell{Low-light}},
        column type=c,
        assign cell content/.code={%
          \pgfmathparse{##1<0.03}%
          \ifnum\pgfmathresult=1
            \pgfkeyssetvalue{/pgfplots/table/@cell content}%
              {$\mathbf{\pgfmathprintnumber[fixed,precision=2,zerofill]{##1}}$}%
          \else
            \pgfkeyssetvalue{/pgfplots/table/@cell content}%
              {$\pgfmathprintnumber[fixed,precision=2,zerofill]{##1}$}%
          \fi
        },
      },%
      columns/{normal-light}/.style={
        column name={\makecell{Office-light}},
        column type=c,
        assign cell content/.code={%
          \pgfmathparse{##1<0.03}%
          \ifnum\pgfmathresult=1
            \pgfkeyssetvalue{/pgfplots/table/@cell content}%
              {$\mathbf{\pgfmathprintnumber[fixed,precision=2,zerofill]{##1}}$}%
          \else
            \pgfkeyssetvalue{/pgfplots/table/@cell content}%
              {$\pgfmathprintnumber[fixed,precision=2,zerofill]{##1}$}%
          \fi
        },
      },%
      columns/{auto-capture}/.style={
        column name={\makecell{Auto-capture}},
        column type=c,
        assign cell content/.code={%
          \pgfmathparse{##1<0.03}%
          \ifnum\pgfmathresult=1
            \pgfkeyssetvalue{/pgfplots/table/@cell content}%
              {$\mathbf{\pgfmathprintnumber[fixed,precision=2,zerofill]{##1}}$}%
          \else
            \pgfkeyssetvalue{/pgfplots/table/@cell content}%
              {$\pgfmathprintnumber[fixed,precision=2,zerofill]{##1}$}%
          \fi
        },
      },%
    ]\bpcerstats
  \endgroup
\end{table}%
\begin{figure}[t]
  \centering
   \includegraphics[width=1.0\linewidth,
                   trim=2mm 12mm 5mm 14mm, %
                   clip]{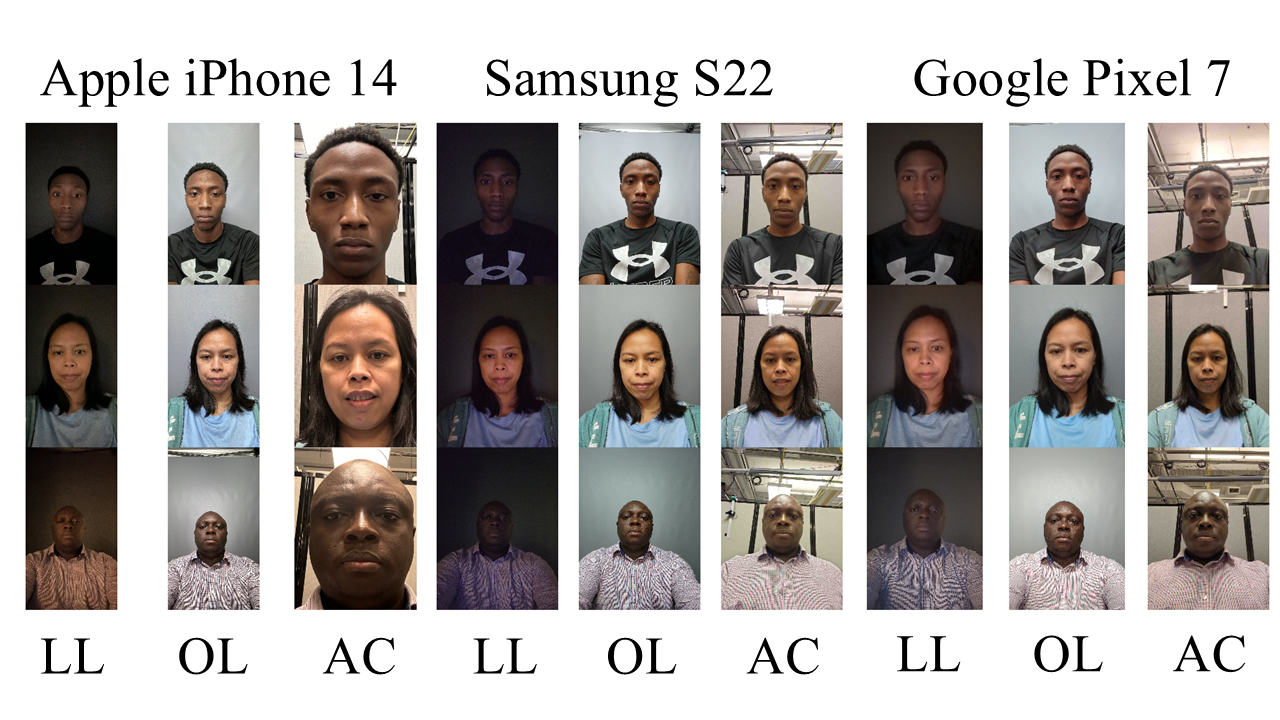}
   \caption{Example images from the three different smartphones and capture scenarios. \textbf{LL}: Low-light capture. \textbf{OL}: Office-light capture. \textbf{AC}: Auto-capture. Differences in face-cropping are due to auto-capture implementation differences between Apple Vision SDK (Apple iPhone) and Google ML kit (Samsung and Google devices). Depicted volunteers gave permission show images in research publications.}
   \label{fig:sample_captures}
\end{figure}%
RIV systems generally consist of three core subsystems: identity document validation, face recognition (for matching selfies to photo identity documents), and presentation attack detection (PAD). Identity document validation involves verifying the authenticity of an identity document by analyzing an image of the document submitted by the user. Face recognition ensures that the individual presenting the document is the same person depicted in the document's photograph. PAD serves as a critical safeguard by detecting attempts to manipulate the biometric authentication process. Because RIV systems are remote by definition, there is an increased opportunity for subversive users to present artifacts such as printed images or masks of another person's biometric sample. By detecting these presentation attacks, PAD helps ensure the integrity and reliability of biometric authentication, making remote identity validation systems more resilient to fraud~\cite{ISOIEC30107-1-2023}. 

In remote identity validation, PAD techniques can be categorized into two types: active and passive~\cite{ngan_face_2023}. Passive PAD techniques validate previously acquired biometric samples, such as images or videos, while active PAD techniques require real-time user interaction with devices, including tablets or smartphones, such as blinking or head movement. Designers of PAD systems must carefully tune their process to provide the security necessary for the application while still efficiently facilitating bona fide users.

PAD subsystem testing is part of broader biometric technology testing which typically involves three main types of evaluation: technology tests, scenario tests, and operational tests. Technology testing evaluates algorithms using a fixed corpus of biometric samples, often collected in advance, and is cost-effective, repeatable, and efficient. However, because this testing is done on a static corpus and under controlled conditions, it may not reflect real-world variables found in actual deployment environments. Scenario and operational testing, in contrast, are conducted in environments that better simulate or directly represent real-world applications, capturing new corpora of images reflecting critical nuances such as device variability, user behavior, and environmental factors, albeit with increased cost and reduced repeatability~\cite{ISOIEC19795-1-2021}.

In remote identity validation scenarios, automated PAD algorithms must operate effectively across diverse real-world environments where capture conditions (e.g., lighting) cannot be controlled. To assess the robustness of PAD under these conditions, this paper examines how low-light environments and automated image capture effects the ability of passive PAD subsystems to facilitate bona fide users. To this end, we conducted a scenario test with a demographically varied group of paid volunteers and evaluate multiple commercial PAD algorithms across three distinct scenarios. The first scenario serves as a baseline, in which PAD algorithms processed images captured under office-lighting with staff-guided image acquisition and quality review. The second scenario examines the impact of low-light conditions on PAD performance, highlighting the challenges that arise when systems are used in non-optimal environments. The third scenario evaluates passive PAD performance using an automated “selfie” capture process reflective of common remote identity validation workflows.

By analyzing PAD performance across these scenarios, this study (1) provides an estimate of facilitation performance of state-of-the-art commercial PAD subsystems in office lighting, low-light, and self-capture scenarios when used with three different smartphone models, (2) demonstrates a model for scenario testing of passive pad algorithms, and (3) identifies that many commercial PAD systems experience a significant decline in performance when utilized in low-light or automated capture scenarios.

\section{Background}
\label{sec:background}%
Passive PAD is commonly evaluated with technology tests relying on publicly available datasets. The NUAA Imposter Database (2010) comprises still images comparing real faces with printed photo attacks for 15 subjects~\cite{tan_face_2010}. Idiap’s Print-Attack~\cite{anjos_counter_2011} and Replay-Attack~\cite{Chingovska_lbp_2012} datasets (2011–2012) introduced varied lighting conditions and replay attacks. CASIA-FASD (2012) provided additional variations in camera quality, ranging from low to high~\cite{Zhang_antispoofing_2012}. MSU Mobile Face Spoofing (2015) emphasized smartphone replay and print attacks~\cite{wen_distortion_2015}, while Replay-Mobile (2016) incorporated diverse indoor and outdoor lighting conditions~\cite{costa_replay_2016}.

The OULU-NPU dataset (2017) systematically evaluated algorithm generalization across different illuminations, camera sensors, and attack devices~\cite{boulkenafet_database_2017}. SiW (2018) highlighted high-resolution videos under varied environmental conditions and protocols designed to test unseen attack media~\cite{liu_face_2018}. The 3DMAD dataset (2013) pioneered realistic 3D mask attacks~\cite{erdogmus_kinect_2013}, and the HKBU Mask datasets (2016, 2018) extended this approach by including multiple mask materials and varied conditions~\cite{liu_antispoof_2016, liu_3d_2016, liu_mask_2018}. WMCA (2020)~\cite{george_pad_2020} and HiFiMask (2021) further increased realism and challenge in mask-based attacks~\cite{liu_mask_2021}. Recent datasets, CelebA-Spoof (2020)~\cite{zhang_celeba_2020} and SOTERIA Face PAD (2024), specifically target extensive diversity and low-light conditions~\cite{ramoly_novel_2024}.

While these publicly available datasets are important for developing PAD technology and providing common benchmarks, their utility for evaluating commercial RIV systems is limited. Algorithm developers can incorporate these datasets into their training processes. Testing using training data results in greater performance relative to testing using “unseen” operational data.  Evaluations of commercial RIV systems therefore require the collection of additional data that simulates operational conditions or the use of carefully sequestered datasets not available during system development.

Other PAD evaluations have utilized sequestered datasets to ensure evaluation integrity and validity. The Intelligence Advanced Research Projects Activity ODIN program (2016–2021) developed and assessed PAD algorithms using sequestered datasets~\cite{iarpa-odin}, but did not specifically study the RIV use-case. In 2023, the National Institute of Standards and Technology conducted its Facial Analysis Technology Evaluation, analyzing commercial passive PAD algorithms, but did not investigate the effects of auto-capture or lighting on the performance~\cite{ngan_face_2023}. The Center for Identification Technology Research conducted operational tests of remote identity verification providers, including PAD capabilities in 2024, but focused on analyzing the differential performance between demographic groups~\cite{fatima_large-scale_2024}. Additionally, the Department of Homeland Security's Science and Technology Directorate (DHS S\&T) Remote Identity Validation Technology Demonstration (RIVTD) evaluated passive PAD algorithms using new sequestered datasets in 2024, but only for well-lit imagery collected using staff supervision~\cite{Howard2025RIVFramework}.

These prior evaluations have contributed meaningfully to PAD research. However, the effects of capture conditions on the performance of commercial PAD technologies remains underexplored despite being of high relevance to successful RIV system deployments.

\section{Experimental Design}
\label{sec:expdesign}%
\subsection{Overview}
\label{sec:expdesign:overview}
\begin{figure}[t]
  \centering
   \includegraphics[width=1.0\linewidth,
                   trim=0mm 2.6mm 1.9mm 3.9mm, %
                   clip]{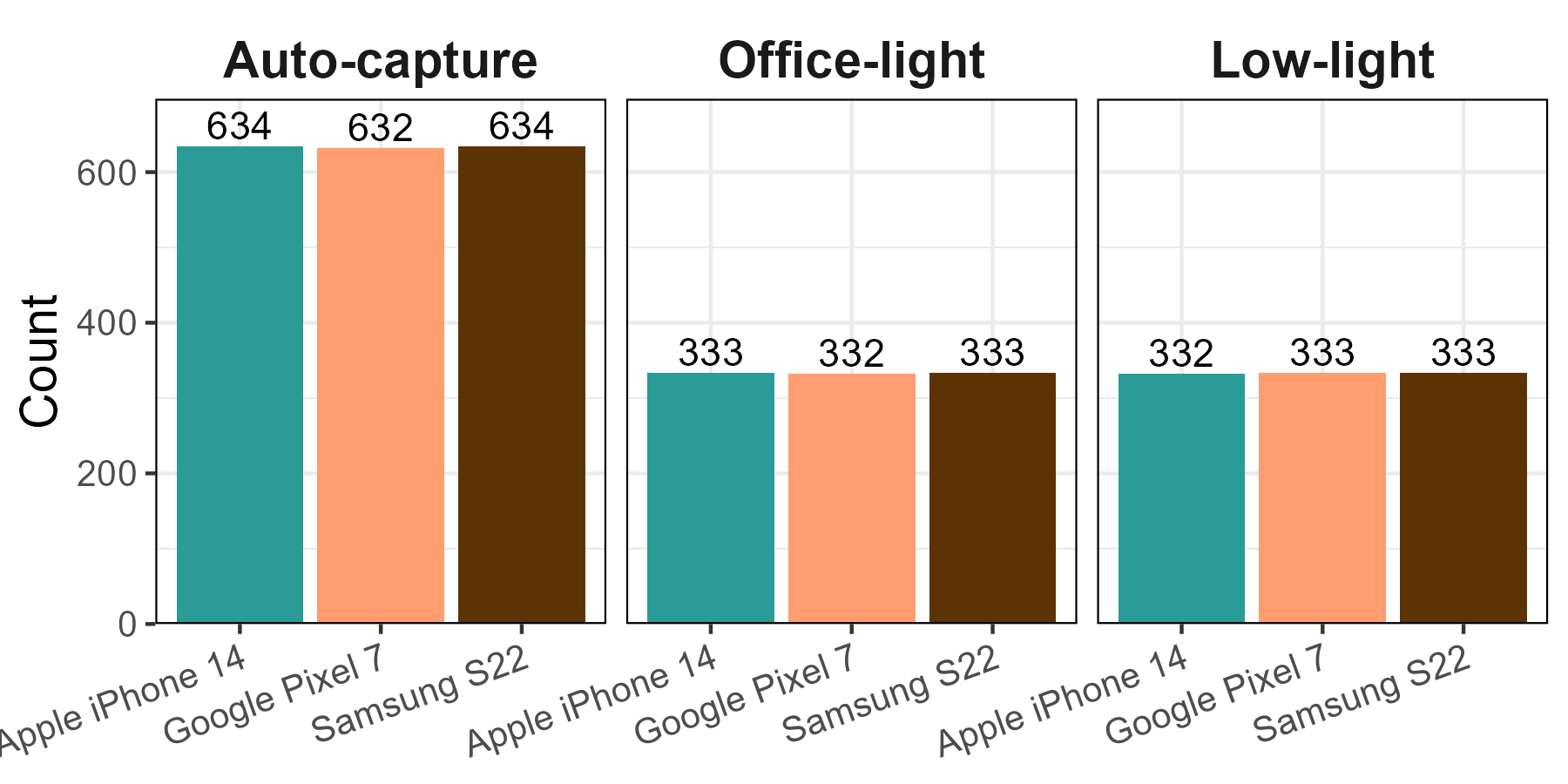}
   \caption{The counts of image samples in each of the scenarios for each device. One image sample per volunteer in each smartphone-scenario bin.}
   \label{fig:count_samples}
\end{figure}

\begin{figure}[t]
  \centering
   \includegraphics[width=1.0\linewidth,
                   trim=1.5mm 2.6mm 1.9mm 3.9mm, %
                   clip]{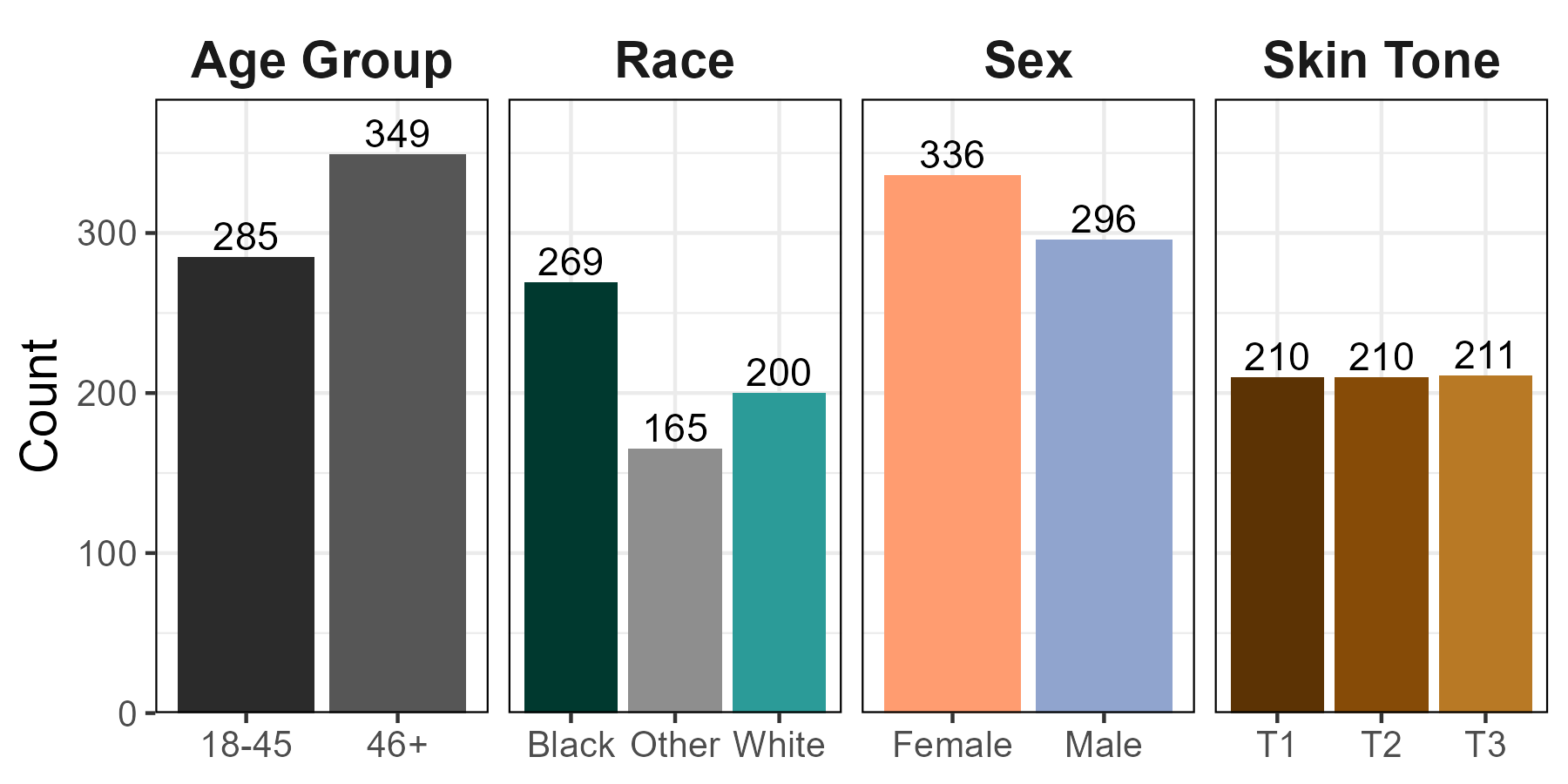}
   \caption{Counts of volunteers by demographic category. T1, T2, T3 are skin tone lightness tertiles of the volunteers.}
   \label{fig:count_volunteers}
\end{figure}

This study tested the isolated influence of low-light and automated capture on the performance of passive PAD systems using a new dataset of images collected as part of a scenario test.  A total of 634 paid volunteers participated in the test in September of 2024. All subjects consented to participate in the study under an established Institutional Review Board (IRB) protocol. Each volunteer completed up to three image-capture scenarios.  All volunteers participated in the auto-capture scenario and approximately half of the group participated in two controlled lighting capture scenarios (Figure~\ref{fig:count_samples}). Images were collected on three currently available smartphones: Apple iPhone 14, Google Pixel 7, and Samsung Galaxy S22. Each volunteer performed a capture on each device. Volunteers were asked to self-report their age, race, and gender at the time of data collection. Self-reported genders were mapped to sex in this report. Skin tone was measured using calibrated colormeter (DSM III, Cortex Technology) and grouped into three tertiles based on the L* component of the average CIELAB reading between left and right temples (T1: \TOneL-\TOneH, T2: \TTwoL-\TTwoH, T3: \TThrL-\TThrH). Some volunteers did not self-report their gender or opted out of skin tone measurement. Demographics of the test population are shown in Figure \ref{fig:count_volunteers}.%

Example images captured in each scenario are shown in Figure \ref{fig:sample_captures}. The images were grouped into nine datasets.   Performance of passive PAD systems was evaluated separately using images in each dataset to analyze how environmental and procedural changes affected performance.%
\subsection{Data Capture Procedure}
\label{sec:expdesign:data_capture_procedure}
\begin{figure}[t]
  \centering
   \includegraphics[width=1.0\linewidth,
                   trim=0mm 0mm 66mm 71mm, %
                   clip]{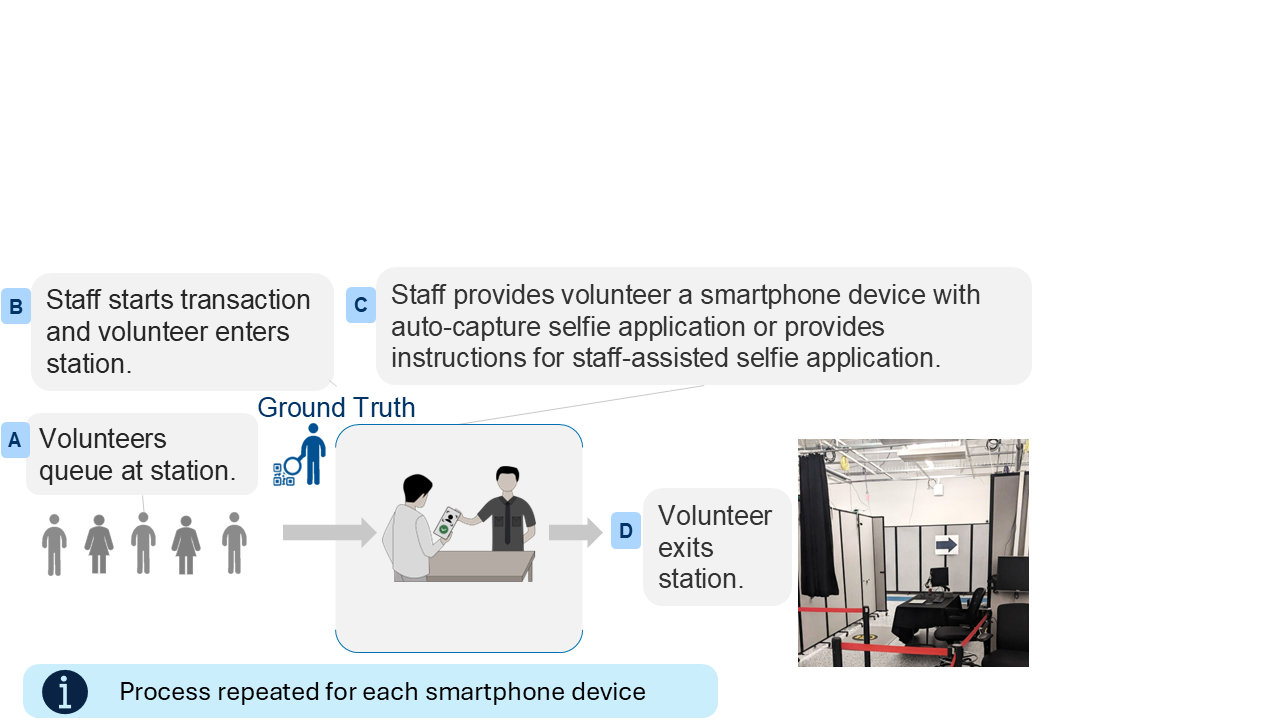}
   \caption{Data capture procedure.}
   \label{fig:station_flow}
\end{figure}

The data capture procedure was designed to isolate the tested factors and ensure reliable association between the volunteers’ identities and the acquired data (Figure \ref{fig:station_flow}). Volunteers first queued at a data capture station, where a staff member linked the volunteers’ ground truth identity with a transaction. The volunteer operated an application on one of the smartphones and the application saved the collected images. The volunteer then exited the station. Three different procedures were implemented for the different scenarios (office-light, low-light, and auto-capture)%

In the controlled lighting capture scenarios (office-light and low-light), image capture was guided by a staff member and performed within an enclosure with controlled lighting.  The staff member verbally guided the volunteer through the capture process. The staff member ensured the proper alignment of the volunteer’s face within the field of view, provided instructions as needed, and manually reviewed the collected data for completeness. Automated face quality checks were not used in this process. If necessary, the staff member requested another capture from the volunteer to maintain adequate quality.

Lighting within the enclosure was configured to have consistent illuminance using a handheld illuminance meter placed horizontally at the expected location of the volunteer.  The office light configuration had an illuminance of 600-650 lux, a color temperature of 6100K-6250K, and high-frontal lighting orientation. The low-light configuration had an illuminance of 0-20 lux.  

\begin{figure}[t]
  \centering
   \includegraphics[width=1.0\linewidth,
                   trim=0mm 0.5mm 120mm 111mm, %
                   clip]{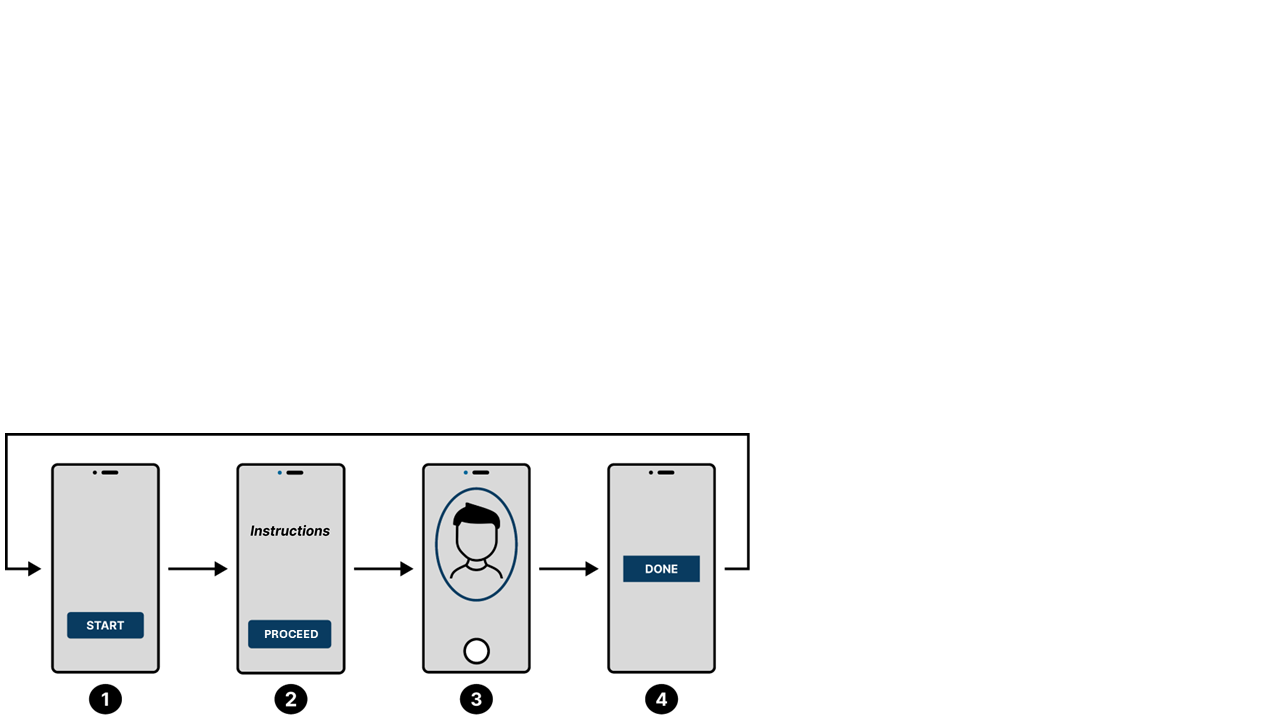}
   \caption{The process flow requirement for the automated capture application. Actual application screens not shown.}
   \label{fig:app_flow}
\end{figure}

The auto-capture scenario simulated a fully automated RIV use-case. The volunteer used an application on each smartphone to automatically capture a selfie image. As illustrated in Figure \ref{fig:app_flow}, the volunteer followed on-screen instructions to complete the capture process independently without any guidance from test staff. The application performed face detection and quality assessment automatically, ensuring a face was centered within the field of view and of an appropriate size.   The capture application was representative of an operational identity verification system.  The system used Apple Vision SDK on iOS and the Google ML Kit on the Android smartphones for face detection and quality checks, introducing differences in face cropping in the collected images. Illuminance in the auto-capture scenario was 600-650 lux, with a warmer color temperature of 3200K-3350K and a more frontal lighting direction relative to the office-light controlled captures.%

\subsection{Passive PAD Systems}
\label{sec:expdesign:passive_pad_systems}
A total of nine passive PAD subsystems, each from a different commercial vendor, were evaluated using datasets consisting of three different smartphones and three different capture scenarios. PAD subsystems were chosen for this study to represent the state-of-the-art in commercial offerings. All algorithms had no specific pose or expression requirements, claimed the ability to detect highly sophisticated presentation attacks, and, except for one, had active operational deployments as of 2024.%
\subsection{Metrics}
\label{sec:expdesign:metrics}
The focus of this study is the facilitation performance of commercial passive PAD systems.  The bona fide presentation classification error rate (BPCER) quantifies the proportion of bona fide presentations misclassified as attacks. Prior to this evaluation, the PAD systems were configured by their commercial vendors to target a 1\% BPCER. Under a failure is suspicious policy, system non-responses contributed to BPCER.

For each subsystem, the maximum BPCER among the three tested smartphones (BPCER\textsubscript{max}) was used to provide a conservative estimate of system performance.  BPCER\textsubscript{max} was compared to a facilitation benchmark of 3\% following the convention set by DHS S\&T in the RIVTD~\cite{Howard2025RIVFramework}.  According to this convention, PAD systems with measured BPCER\textsubscript{max} below 3\% meet the facilitation benchmark. 

To assess the general effect of each scenario (low-light, office-light, and auto-capture) on the commercial PAD systems, we measured the proportion of subsystems that met the benchmark under each scenario (Table \ref{tab:bpcer_max}). This approach provided a comparative analysis of how different conditions influenced the facilitation offered by commercial passive PAD systems.%
\subsection{Statistical Analysis}
\label{sec:expdesign:statistical_analysis}
Error counts were analyzed with a hierarchical binomial‐logistic mixed model (Equation \ref{eq:glmm}) fitted by maximum-likelihood using the lme4 R package (glmer, Laplace approximation)~\cite{bates2015lme4,r_core_team_r_2024}, modeling the probability of error for subsystem $j$, device $k$, scenario $s$ ($p_{j,k,s}$).  Scenario was the only fixed predictor, coded with three levels: office-light as the reference ($\beta_0$), low-light ($\beta_{low}$), and auto-capture ($\beta_{auto}$). The resulting contrasts quantify, for each alternate capture scenario, the log-odds ratio compared with office-light. To account for the hierarchical sampling design, multiple devices were nested within each tested subsystem, and we specified independent random intercepts for subsystems ($\upsilon_j$) and for smartphones within subsystems ($\nu_{j,k}$). This structure allows baseline error propensities to vary both between subsystems and among smartphones for the same subsystem, thereby controlling for clustering in the data. Model convergence was checked with the BOBYQA optimizer~\cite{powell2009bobyqa} and default diagnostic criteria. Statistical inference on fixed effects used Wald z-tests~\cite{agresti2013cda}.%
\begin{equation}
        \operatorname{logit} p_{j,k,s}
        = \beta_{0} + \upsilon_{j} + \nu_{j,k}
        + \left\{
        \begin{array}{l}
        0\\[2pt]
        \beta_{\text{low}}\\[2pt]
        \beta_{\text{auto}}
        \end{array}
        \right.
      \label{eq:glmm}
\end{equation}

\section{Results}
\label{sec:results}%
Table \ref{tab:bpcer_max} presents the BPCER\textsubscript{max} values for the different tested vendor PAD subsystems under each scenario. The office-light scenario yielded the highest success rate, with 78\% of systems meeting the 3\% facilitation benchmark. The auto-capture scenario followed with a 44\% benchmark success rate, while the low-light scenario had the lowest success rate at 11\%. Three PAD subsystems (C, E, and I) were partially successful by meeting the benchmark in both the default and auto-capture scenarios, but not low-light. Despite being tested in a controlled and well-lit environment, two subsystems (B and F) failed to meet the benchmark even in the office-light scenario, highlighting the challenges of achieving low BPCER for PAD.%
\begin{table}[t]
  \centering
  \caption{The BPCER\textsubscript{max} values for the different PAD subsystems for low-light, office-light, and auto-capture scenarios. Bold values indicate when the 3\% threshold has been successfully met. The final row provides the total percentage of PAD subsystems able to remain below the 3\% benchmark for each scenario.}
  \label{tab:bpcer_max}
  \begingroup
    \setlength{\tabcolsep}{4.2pt}
    \begin{filecontents*}{./data/bpcer_max_summary_riv_usability_no_gots_met_benchmark.csv}
alias,low-light,normal-light,auto-capture
A,0.0512048192771084,0.012012012012012,0.0883280757097792
B,0.309309309309309,0.441441441441441,0.239747634069401
C,0.659638554216867,0.00301204819277108,0.00158227848101266
D,0.0690690690690691,0,0.209779179810726
E,0.246987951807229,0,0.0236593059936909
F,0.498498498498498,0.045045045045045,0.389589905362776
G,0.018018018018018,0,0
H,0.0630630630630631,0.021021021021021,0.0772870662460568
I,0.045045045045045,0.003003003003003,0.00158227848101266
Met Benchmark,0.111111111111111,0.777777777777778,0.444444444444444
\end{filecontents*}
\pgfplotstabletypeset[
      display columns/0/.style={
        column name={PAD Subsystem},
        column type=l
      },%
      display columns/1/.style={
        column name={Low-light},
        column type=c,
        assign cell content/.code={%
          \pgfmathparse{##1<0.03}%
          \ifnum\pgfmathresult=1
            \pgfkeyssetvalue{/pgfplots/table/@cell content}%
              {$\mathbf{\pgfmathprintnumber[fixed,precision=2,zerofill]{##1}}$}%
          \else
            \pgfkeyssetvalue{/pgfplots/table/@cell content}%
              {$\pgfmathprintnumber[fixed,precision=2,zerofill]{##1}$}%
          \fi
        },
      },%
      display columns/2/.style={
        column name={Office-light},
        column type=c,
        assign cell content/.code={%
          \pgfmathparse{##1<0.03}%
          \ifnum\pgfmathresult=1
            \pgfkeyssetvalue{/pgfplots/table/@cell content}%
              {$\mathbf{\pgfmathprintnumber[fixed,precision=2,zerofill]{##1}}$}%
          \else
            \pgfkeyssetvalue{/pgfplots/table/@cell content}%
              {$\pgfmathprintnumber[fixed,precision=2,zerofill]{##1}$}%
          \fi
        },
      },%
      display columns/3/.style={
        column name={Auto-capture},
        column type=c,
        assign cell content/.code={%
          \pgfmathparse{##1<0.03}%
          \ifnum\pgfmathresult=1
            \pgfkeyssetvalue{/pgfplots/table/@cell content}%
              {$\mathbf{\pgfmathprintnumber[fixed,precision=2,zerofill]{##1}}$}%
          \else
            \pgfkeyssetvalue{/pgfplots/table/@cell content}%
              {$\pgfmathprintnumber[fixed,precision=2,zerofill]{##1}$}%
          \fi
        },
      },%
      every row no 9/.style={before row=\midrule}, %
    ]{./data/bpcer_max_summary_riv_usability_no_gots_met_benchmark.csv}
  \endgroup
\end{table}%
\setlength{\parindent}{1.5em} %
\setlength{\parskip}{0pt}

Table \ref{tab:bpcer} provides the BPCER for each combination of PAD subsystem and vendor smartphone. Average, standard deviation, and median values for each capture scenario are in Table \ref{tab:bpcerstats}. The average PAD system performance in all three scenarios was above 3\% BPCER due to high error rates on some systems.  The median PAD system performance in the office-light and auto-capture scenarios was below the 3\% BPCER benchmark.%
\pgfplotstableread{./data/bpcer_summary_riv_usability.csv}\bpcer
\begin{table}[t]
  \centering
  \caption{The detailed BPCER values for the different PAD subsystem, smartphone, and scenario combinations. Bold values indicate when the 3\% threshold has been successfully met.}
  \label{tab:bpcer}
  \begingroup
    \setlength{\tabcolsep}{2.725pt}
    \pgfplotstabletypeset[
      columns/{alias}/.style={
        column name={\makecell[l]{PAD \\Subsystem}},
        column type=l
      },%
      columns/{device}/.style={
        column name={Smartphone},
        column type=l
      },%
      columns/{low-light}/.style={
        column name={\makecell{Low-\\light}},
        column type=c,
        assign cell content/.code={%
          \pgfmathparse{##1<0.03}%
          \ifnum\pgfmathresult=1
            \pgfkeyssetvalue{/pgfplots/table/@cell content}%
              {$\mathbf{\pgfmathprintnumber[fixed,precision=2,zerofill]{##1}}$}%
          \else
            \pgfkeyssetvalue{/pgfplots/table/@cell content}%
              {$\pgfmathprintnumber[fixed,precision=2,zerofill]{##1}$}%
          \fi
        },
      },%
      columns/{normal-light}/.style={
        column name={\makecell{Office-\\light}},
        column type=c,
        assign cell content/.code={%
          \pgfmathparse{##1<0.03}%
          \ifnum\pgfmathresult=1
            \pgfkeyssetvalue{/pgfplots/table/@cell content}%
              {$\mathbf{\pgfmathprintnumber[fixed,precision=2,zerofill]{##1}}$}%
          \else
            \pgfkeyssetvalue{/pgfplots/table/@cell content}%
              {$\pgfmathprintnumber[fixed,precision=2,zerofill]{##1}$}%
          \fi
        },
      },%
      columns/{auto-capture}/.style={
        column name={\makecell{Auto-\\capture}},
        column type=c,
        assign cell content/.code={%
          \pgfmathparse{##1<0.03}%
          \ifnum\pgfmathresult=1
            \pgfkeyssetvalue{/pgfplots/table/@cell content}%
              {$\mathbf{\pgfmathprintnumber[fixed,precision=2,zerofill]{##1}}$}%
          \else
            \pgfkeyssetvalue{/pgfplots/table/@cell content}%
              {$\pgfmathprintnumber[fixed,precision=2,zerofill]{##1}$}%
          \fi
        },
      },%
      every row no 3/.style={before row=\midrule}, %
      every row no 6/.style={before row=\midrule},
      every row no 9/.style={before row=\midrule},
      every row no 12/.style={before row=\midrule},
      every row no 15/.style={before row=\midrule},
      every row no 18/.style={before row=\midrule},
      every row no 21/.style={before row=\midrule},
      every row no 24/.style={before row=\midrule},
    ]\bpcer
  \endgroup
\end{table}%

\begin{figure}[t]
  \centering
   \includegraphics[width=1.0\linewidth,
                   trim=13mm 25mm 13mm 1mm, %
                   clip]{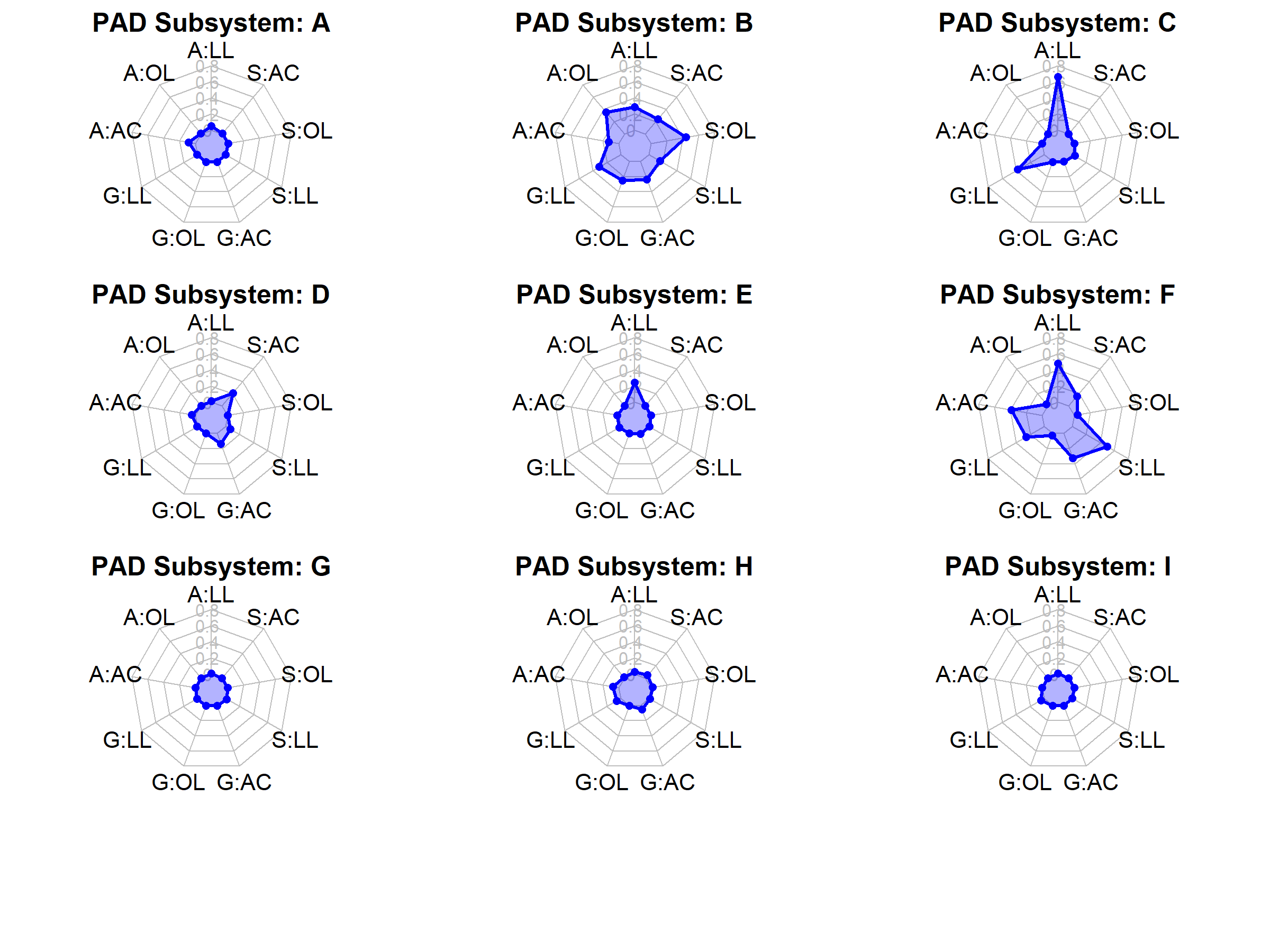}
   \caption{Visualization of PAD system robustness.  Each subplot shows BPCER measurements for a single PAD system.  Each axis represents the BPCER measurement for a specific smartphone in a single scenario. Axes are labeled as 'S:CC' where S is the first letter of the smartphone model (A: Apple iPhone 14, G: Google Pixel 7, S: Samsung Galaxy S22), and CC represents the scenario (\textbf{LL}=Low-light, \textbf{OL}=Office-light, \textbf{AC}=Auto-capture). Shapes closer to the center indicate lower BPCERs,  Outward extensions along some axes indicate higher error rates and weaker performance in specific capture conditions.}
   \label{fig:radar}
\end{figure}

The impact of the different scenarios was not uniform across all vendor subsystems. Notably, PAD G met the 3\% BPCER benchmark across all smartphones and scenarios, while no other PAD algorithm achieved the benchmark under low-light conditions for all three smartphones. Additionally, four PAD subsystems (A, G, H, and I) were robust across capture conditions, as shown by their relatively consistent performance across smartphone–scenario combinations, visualized by the radial uniformity in Figure \ref{fig:radar}. In contrast, subsystems C, D, E, and F are fragile to capture conditions, as visualized by radial deviations in the same figure.%

To quantify how the low-light and auto-capture scenarios influence the performance of a generic passive PAD subsystem relative to office-light, we analyzed the error-count data from all smartphones and PAD subsystems with a hierarchical binomial model nesting smartphones within each PAD subsystem. The fit parameters (Table \ref{tab:model-params}) shows that low-light increases the odds of an error by about four-fold relative to office-light $(p < 0.001)$, while the auto-capture workflow roughly doubles those odds $(p < 0.001)$. The model allows these effects to be translated to the probability scale. For example, the model predicts that a PAD system with a BPCER of 1.4\% in the office-light scenario (the grand mean intercept in office-light scenario) would have a BPCER of 2.8\% under auto-capture and a BPCER of 5.4\% in low-light. Auto-capture therefore pushes the mean system with typical performance to the edge of the 3\% facilitation benchmark, whereas low-light clearly breaches it. These penalties are large and statistically significant, underscoring low-light as the greater practical risk and auto-capture as a near-threshold concern. The model also provides us insight into the large effects of vendor choice and smartphone model on BPCER. One standard deviation of vendor performance from the grand mean represents a substantial range in BPCER: 0.30\% to 6.30\%. Additionally one standard deviation of smartphone model performance represents the BPCER range 0.50\% to 3.9\%. The model implies that there is almost as much variation in performance across smartphones as there is across vendors.%
\begin{table}[t]
  \centering
  \begin{tabular}{lcc}
    \toprule
    \textbf{Parameter} & \textbf{Estimate} & \textbf{Probability} \\
    \midrule
    $\beta_{0}$                              & $-4.27~(p < 0.001)$           & $0.014$ \\
    $\upsilon_{j}\!: N(\mu,\sigma^{2})$          & $N(0,\,2.46)$     & \textemdash{} \\
    $\nu_{j,k}\!: N(\mu,\sigma^{2})$         & $N(0,\,1.13)$     & \textemdash{} \\
    $\beta_{\text{low}}$                    & $1.40~(p < 0.001)$            & $0.054$ \\
    $\beta_{\text{auto}}$                   & $0.70~(p < 0.001)$            & $0.028$ \\
    \bottomrule
  \end{tabular}
  \caption{Fit parameters of the hierarchical binomial‐logistic mixed model. $\beta_0$: fixed intercept, $\upsilon_j$: subsystem random intercept for subsystem $j$, $\nu_{j,k}$: device-within-subsystem random intercept for device $k$ inside subsystem $j$, $\beta_{low}$: the fixed effect added when the scenario is low-light, $\beta_{auto}$: the fixed effect added when the scenario is auto-capture.}
  \label{tab:model-params}
\end{table}

\section{Discussion}
\label{sec:discussion}%

This study examined the ability of commercial PAD systems to maintain robust facilitation across variations in capture conditions.  The results demonstrated that ambient lighting, capture method, and smartphone model can cause significant variations in BPCER on many of the tested systems.  Indeed, only one of the nine tested systems was both robust across capture conditions and consistently maintained error rates below 3\%.  Lack of robustness in PAD system performance can degrade the user experience of remote identity systems.

Testing PAD systems in low-light conditions revealed an increase in error rates, underscoring a realistic scenario in which RIV systems must function independently. Since users often engage with these systems on their own, interactions can occur at night or in settings with inadequate lighting. Additionally, interactions with RIV systems require a user interface and auto capture pipeline, which is rarely included during system testing. In our testing only 44\% of systems met the BPCER\textsubscript{max} benchmark with the unsupervised auto capture pipeline compared to the 78\% of systems capable of meeting the benchmark when using manually collected and reviewed imagery. While evaluations conducted in ideal lighting conditions and with manually collected imagery provide valuable benchmarks, they may overestimate real-world PAD performance. 

This study also observed significant variation in performance between different vendor subsystems and smartphone models, ranging from 0\% to 44\% BPCER in the office-light scenario. When looking at the performance variation due to smartphones within individual vendor subsystems, the biggest variabilities are apparent in the low-light scenario, with subsystems D, E, H, only able to meet the 3\% benchmark on certain devices. This highlights the importance of vendor selection on bona fide user facilitation, reinforcing the importance of independent evaluations of RIV system performance to ensure low error rates across capture conditions. It also suggests that smartphone model is a significant source of performance variability for certain vendors, possibly exacerbated by variations in lighting. Often overlooked as a factor during testing, the smartphone model used for RIV can be among thousands of different potential options, each with distinct cameras, camera configurations, and processing pipelines. As RIV systems are expected to work broadly, future research should explore this aspect to advance understanding into how variations in smartphone hardware effect PAD effectiveness.

When placed in real-world scenarios, lack of robustness in PAD system performance can lead system owners to prioritize high facilitation at the expense of security. As bona fide presentations dominate in operational settings, system owners may feel pressured to reconfigure systems to facilitate users of RIV systems at the expense of detecting presentation attacks.  Configuring fragile PAD systems to maintain a low BPCER can lead to an overall reduced ability to detect attacks even when capture conditions are favorable.  Selecting robust PAD systems is therefore crucial to maintaining both facilitation and security in RIV system deployments.

\section{Conclusion}
\label{sec:conclusion}%
This study provides an in-depth analysis of how environmental and procedural changes affect the robustness of PAD systems in RIV. In this analysis we provided an estimate of facilitation performance of state-of-the-art commercial PAD systems in office lighting, low-light, and self-capture scenarios across nine different vendor subsystems and three different smartphones. In doing so, we demonstrated a model for scenario testing of passive PAD algorithms and identified that many commercial PAD systems experience a decline in mean, median, and model-predicted performance when utilized in low-light or automated capture scenarios. By continuing to test PAD systems in a variety of contexts, RIV technology can continue evolving to provide secure and robust identity verification solutions for real-world applications.

\section{Acknowledgments}
\label{sec:acknowledgments}%
This research was sponsored by the United States Department of Homeland Security’s Science and Technology Directorate on contract number 70RSAT23CB00000003.  Author contributions: ROP conceived the work, designed and performed statistical analyses, wrote and edited the paper and is the corresponding author.  YBS conceived the work, directed the project, contributed to statistical analyses, writing, and editing.  JLT and ARV conceived the work. JJH edited the paper and reviewed statistical methods. Special thanks to SAIC Identity and Data Sciences Laboratory members Cynthia M. Cook for expert review and contributions to the statistical methods and to Rebecca E. Rubin for copy-editing support. Planning and execution of the scenario tests and data collection was performed by the multi-disciplinary staff of the SAIC Identity and Data Sciences Laboratory for the DHS S\&T Biometric and Identity Technology Center.

{
    \small
    \bibliographystyle{ieeenat_fullname}
    \bibliography{main}
}

\end{document}